\documentclass{article}

    \usepackage[final]{project}

\usepackage[utf8]{inputenc} 
\usepackage[T1]{fontenc}    
\usepackage{hyperref}       
\usepackage{url}            
\usepackage{booktabs}       
\usepackage{amsfonts}       
\usepackage{nicefrac}       
\usepackage{microtype}      
\usepackage{caption}
\usepackage{subcaption}
\usepackage[table]{xcolor}
\usepackage{longtable}
\usepackage{multirow}
\usepackage{amsmath}
\usepackage{amsfonts}
\usepackage{amssymb}
\usepackage{float}
\usepackage{graphicx}
\usepackage{lipsum}  
\usepackage{multicol}

\title{Elucidating the Role of Feature Normalization in IJEPA}

\author{%
  Adam Colton \\
  Harmony AI \\
  \texttt{adam@useharmony.com} \\
  \href{https://useharmony.com}{https://useharmony.com}
  \AND
}

\begin{document}

\maketitle 

\begin{abstract}

In the standard image joint embedding predictive architecture (IJEPA), features at the output of the teacher encoder are layer normalized (LN) before serving as a distillation target for the student encoder and predictor. We propose that this feature normalization disrupts the natural energy hierarchy of visual tokens, where high-energy tokens (those with larger L2 norms) encode semantically important image regions. LN forces all features to have identical L2 norms, effectively equalizing their energies and preventing the model from prioritizing semantically rich regions. We find that IJEPA models trained with feature LN exhibit loss maps with significant checkerboard-like artifacts. We propose that feature LN be replaced with a DynTanh activation as the latter better preserves token energies and allows high-energy tokens to greater contribute to the prediction loss. We show that IJEPA trained with feature DynTanh exhibits a longer-tailed loss distribution and fixes the checkerboard artifacts in the loss map. Our empirical results show that our simple modification improves ImageNet linear probe accuracy from 38\% to 42.7\% for ViT-Small and reduces RMSE by 0.08 on NYU Depth V2 monocular depth estimation. These results suggest that preserving natural token energies is crucial for effective self-supervised visual representation learning.
\end{abstract}

\section{Introduction}

IJEPA has emerged as an successful self-supervised framework for learning discriminative features without relying on labeled data \cite{assran2023selfsupervisedlearningimagesjointembedding}. IJEPA models learn to encode image patches into features that are useful for predicting masked regions, a task very similar to masked image modeling (MIM) techniques such as masked autoencoders (MAEs) \cite{he2021maskedautoencodersscalablevision}. However, instead of predicting ground truth pixel values as done in MAE, the model predicts layer normalized features that are obtained from a momentum teacher encoder. This alternate strategy is motivated by the observation that learning to predict latent representations is more effective for encoding discriminative features than learning to predict convincing pixel values \cite{zhou2022ibotimagebertpretraining, grill2020bootstraplatentnewapproach, chen2024deconstructingdenoisingdiffusionmodels}.

We aim to tackle a critical yet underexplored aspect of IJEPA: how the encoder's features are processed before serving as contexts and targets for the prediction model. In the standard IJEPA implementation, hidden states at the output of the student encoder are processed by a LN before being passed to the prediction model. The prediction model is then tasked with predicting masked tokens that are standardized along the hidden dimension. The encoder's LN can be seen as serving two roles. Firstly, it stabilizes the training loss by squashing the magnitude of outliers. Secondly, as we show in our experiments, it reduces the variance of the loss over many samples. 

While LN shows strong empirical results, we argue that it introduces a fundamental limitation. LN dilutes the energy hierarchy by equalizing the statistical properties of every token. This flat energy space conflicts with our goal of obtaining an encoder that can understand which image regions contain more discriminative information. We hypothesize that feature LN reduces the surprise of semantically meaningful tokens and that replacing feature LN with a magnitude preserving DynTanh will improve feature learning. 

We propose a modified IJEPA architecture where LN is replaced by a simple DynTanh activation function. DynTanh provides similar stabilizing effects of LN while preserving vital information about relative token magnitude \cite{zhu2025transformersnormalization}. We validate our modification by comparing IJEPA models trained with LN at the output of the encoder, to IJEPA models trained with DynTanh at the output of the encoder. Across different diverse setups we observe that this simple modification improves downstream classification and dense prediction tasks.

\section{Related Work}

Self-supervised learning (SSL) has become a cornerstone of modern machine learning. This paradigm allows researchers to train deep and powerful models on large scale unlabeled datasets. Visual SSL models fit a surrogate task, learning general purpose features that can later be used for a variety of downstream applications \cite{shwartzziv2023compresscompressselfsupervisedlearning}. Once pretrained on a dataset of images, SSL models can be adapted to guide depth prediction models \cite{yang2024depthanythingunleashingpower}, generative models \cite{yu2025representationalignmentgenerationtraining}, or serve as a frozen visual preprocessor for large language models \cite{fan2025scalinglanguagefreevisualrepresentation}.

A dominant paradigm in visual SSL is the use of Siamese networks for self-distillation. In this framework two networks are used: a student and a teacher. The student is trained to predict the output of the teacher for a different augmentation of the same input. The teacher network is typically not updated by backpropagation. Instead its weights are an exponential moving average (EMA) of the student's weights. EMA has been shown to stabilize training and prevent collapse to a trivial solution \cite{grill2020bootstraplatentnewapproach, chen2024deconstructingdenoisingdiffusionmodels}. This student-teacher dynamic was popularized by works like Bootstrap Your Own Latent (BYOL) \cite{grill2020bootstraplatentnewapproach} and DINO \cite{caron2021emergingpropertiesselfsupervisedvision}.

Within this self-distillation framework, various pretext tasks have been proposed. Early successes came from contrastive methods, which pull augmented views of the same image together in an embedding space while pushing apart views from different images \cite{chen2020simpleframeworkcontrastivelearning}. More recently, masked image modeling (MIM) has emerged as a highly effective alternative. MIM approaches train a model to reconstruct masked or missing parts of an input. While some methods, like Masked Autoencoders (MAE) \cite{he2021maskedautoencodersscalablevision}, reconstruct raw pixel values, joint-embedding predictive architectures (JEPA) perform this task in a latent space. IJEPA \cite{assran2023selfsupervisedlearningimagesjointembedding}, the focus of our work, is a prime example. It learns by predicting the teacher network's representations for masked patches using the student's representations of visible context patches. This encourages the model to learn semantic features rather than high-frequency visual details.

A critical design choice in Siamese networks is how the teacher's features are processed before serving as the distillation target. These post-processing strategies can be broadly grouped into two families.
One family converts features into probability distributions. DINO and its successor DINOv2 \cite{caron2021emergingpropertiesselfsupervisedvision, oquab2024dinov2learningrobustvisual}, as well as iBOT \cite{zhou2022ibotimagebertpretraining}, use a projection head followed by a temperature-scaled softmax. The student is then trained with a cross-entropy loss to match the teacher's soft probability distribution. Masked Siamese Networks (MSN) \cite{assran2022maskedsiamesenetworkslabelefficient} take a similar approach, but generate targets by matching features to a codebook of learnable prototypes.

Another family regresses directly onto continuous feature vectors. BYOL, for instance, projects features onto the unit sphere and minimizes the mean squared error between the student's and teacher's vectors \cite{grill2020bootstraplatentnewapproach}. IJEPA operates similarly but uses Layer Normalization (LN) on the teacher's output features, which are then predicted by the student and predictor using the Smooth L1 loss. We argue that this token-wise normalization, common to both BYOL and IJEPA, forces all tokens to have an equal magnitude, thereby discarding the intrinsic "energy" of a token which may encode its semantic importance. While ablations in BYOL and iBOT confirm that removing their respective normalization or centering steps hurts performance \cite{grill2020bootstraplatentnewapproach, zhou2022ibotimagebertpretraining}, our work is the first to investigate replacing magnitude-destroying normalization with a magnitude-preserving alternative to improve the quality of the learned representations.




\section{Methods}

We describe our modified IJEPA training regime. We make several key modifications that allow us to better utilize our limited compute resources. We use the same core principles as IJEPA, patchifying images and assigning a proportion of patches as either context or targets. Differing from vanilla IJEPA, we leverage the technique introduced by Patch N' Pack \cite{dehghani2023patchnpacknavit} to vary both mask rates and image sizes within a batch, significantly improving training efficiency. Additionally, we do away with the block masking scheme employed by IJEPA, instead using a simple window masking approach. Our modified regime allows us to quickly test architectural modifications and better understand the important role of feature normalization.

\subsection{Image Processing Pipeline}

\begin{figure}
    \centering
    \includegraphics[width=0.5\linewidth]{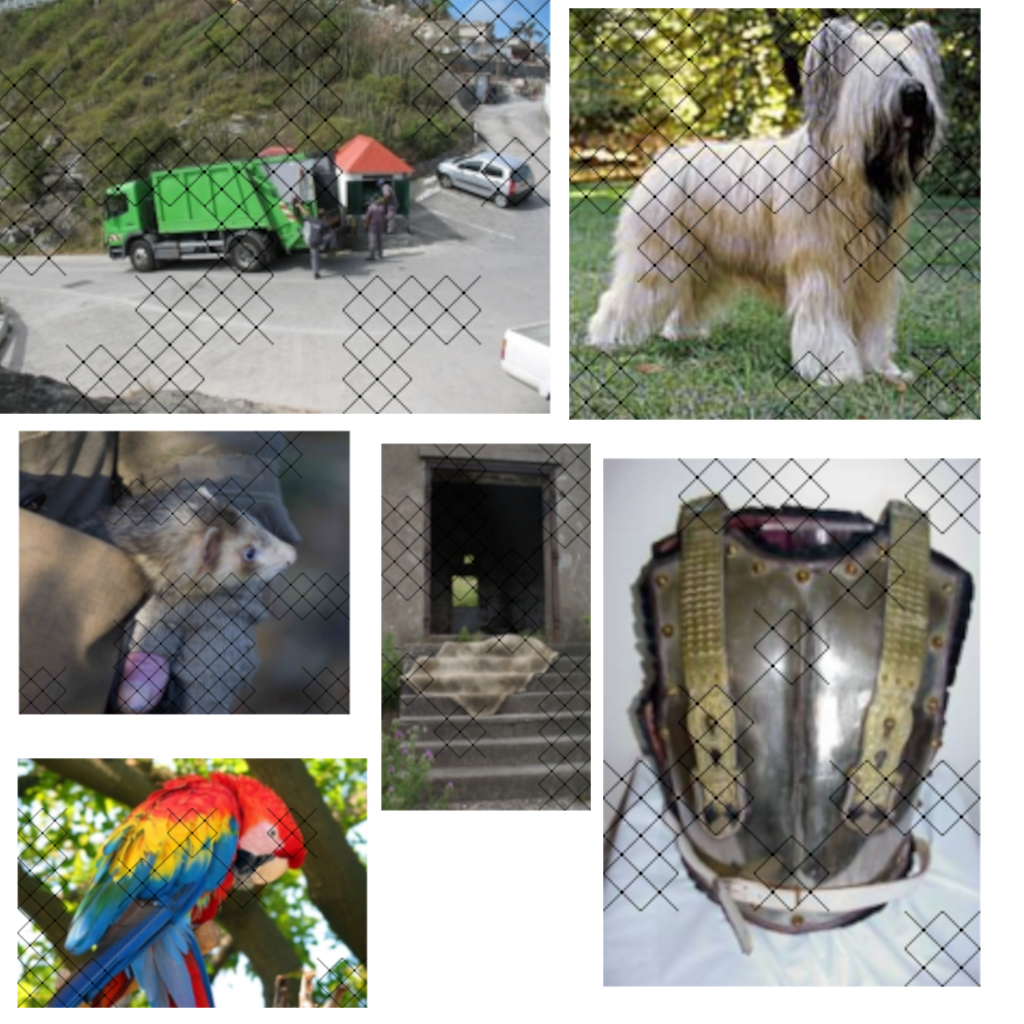}
    \caption{Random samples with mask window size set to two by two patches}
    \centering
    \includegraphics[width=0.5\linewidth]{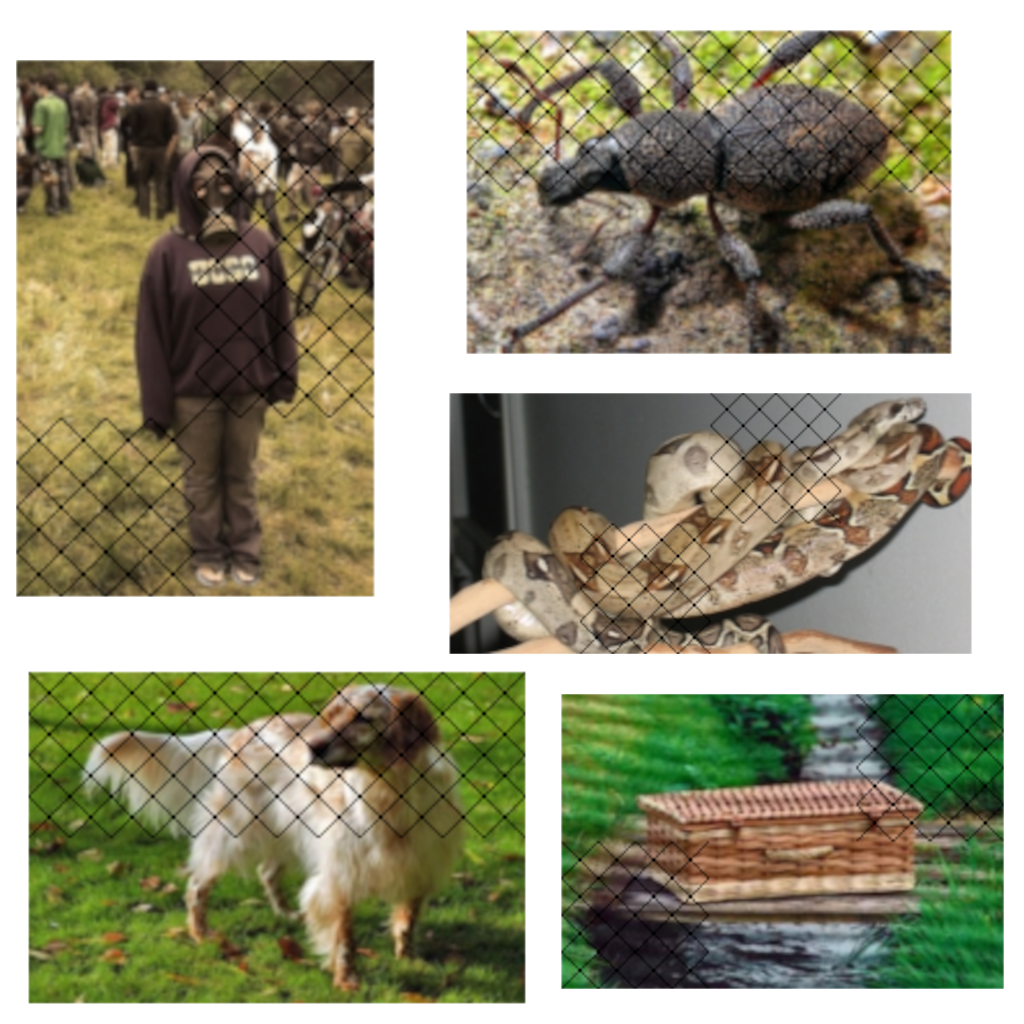}
    \caption{Random samples with mask window size set to four by four patches}
    \label{fig:input-samples}
\end{figure}

Our image processing pipeline consists of the following steps:

\begin{enumerate}

\item \textbf{Dynamic Resolution Scaling}: An input image $\bf y \in \mathbb{R}^{H_{\text{orig}} \times W_{\text{orig}} \times C}$ is downscaled using a random scaling factor $s \sim \mathcal{U}(0.1, 1)$, resulting in an image of size $H \times W \times C$ where $H = s \cdot H_{\text{orig}}$ and $W = s \cdot W_{\text{orig}}$. We enforce the resized height and width to be no smaller than a minimum height and width, and the resized height and width to be divisible by the model's patch size.

\item \textbf{Patch Extraction}: The scaled image is divided into non-overlapping square patches, yielding a set of tokens $y \in \mathbb{R}^{H_p \times W_p \times (P^2 \cdot C)}$, where $P$ is the patch size, and $H_p = H/P$ and $W_p = W/P$ represent the number of patches along height and width dimensions.

\item \textbf{Context-Target Partitioning}: We sample a random context capacity $c \sim \mathcal{U}(0.25, 0.5)$ and assign this proportion of image patches to set $A$, $x_A \in \mathbb{R}^{n \times d_{\text{in}}}$, with the remaining patches assigned to set $B$, $x_B \in \mathbb{R}^{m \times d_{\text{in}}}$. Here, $d_{\text{in}} = P^2 \cdot C$, $n = \lfloor H_p \cdot W_p \cdot c \rfloor$, and $m = H_p \cdot W_p - n$. 

Rather than partitioning image patches at the unit of a single patch, we treat image patches as square windows. This changes the unit of masking from patches into windows. Example outputs of this partitioning can be found in Figure \ref{fig:input-samples}.

\item \textbf{Dual Sequence Packing}: We implement an online dual-stream greedy bin packing algorithm to efficiently batch variable-length sequences. Given a batch size $B$, we allocate two buffers: one of shape $B \times N$ for packing context sequences and another of shape $B \times M$ for packing target sequences, where $N$ and $M$ represent the maximum sequence lengths for context and target, respectively. For each sample, we place its context-target pair in the first available buffer position that can accommodate both sequences simultaneously, ensuring they occupy the same batch index. We track each token's provenance using sample and position identifiers, which are used to compute an attention mask that prevents tokens from different samples from attending to each other. This approach, similar to the packing strategy in \cite{dehghani2023patchnpacknavit}, maximizing computational efficiency by maintaining high occupancy rates in both streams.

\end{enumerate}

\subsection{IJEPA Pretraining}

The core design of our approach maintains the key principles of the original IJEPA framework \cite{assran2023selfsupervisedlearningimagesjointembedding}. A teacher encodes both the context and target regions into features. A predictor uses the student's features to try and predict the teacher's features given only the positions of these target features.

\subsubsection{Formal definition}

After the patch partitioning described in the previous section, the student encoder $f(\cdot)$ processes patches from the context set $A$, producing raw features:
\begin{equation}
\tilde S_x = f(x_A) \in \mathbb{R}^{N \times d}
\end{equation}
where $d$ is the hidden dimension of the model.
Concurrently, the teacher encoder $f'(\cdot)$ processes the complete set of patches (both context and target), yielding raw features:
\begin{equation}
\tilde S_{y_{AB}} = f'(x_A \oplus x_B) \in \mathbb{R}^{(N+M) \times d}
\end{equation}
where $\oplus$ denotes concatenation.

We then take only the tokens originating from the target patches, to obtain our raw target features:

\begin{equation}
\tilde S_y \in \mathbb{R}^{M\times d}
\end{equation}

Both representation undergo layer normalization with learnable affine parameters:
\begin{equation}\label{eq:LN-postproc}
S_x = \text{LN}_{student}(\tilde S_x), S_y = \text{LN}_{
teacher}(\tilde S_y)
\end{equation}

We apply two additional operations:
\begin{enumerate}
\item Batch repetition by factor $r$ to increase effective batch size of the predictor. In our experiments we set $r=4$.
\item Random token dropout both sequences, using a fixed rate $p_{\text{drop}}$ from both $S_x$ and $S_y$. In our experiments we set $p_{\text{drop}}=0.75$
\end{enumerate}

The predictor network $g(\cdot)$ then transforms the student features to predict the teacher's features at target locations:
\begin{equation}
\hat S_y = g(S_x, \mathcal{P}) \in \mathbb{R}^{M \times d}
\end{equation}
where $\mathcal{P}$ represents the positions of target patches.

The training objective minimizes the smooth L1 distance between predicted and actual teacher features.
\begin{equation}
\mathcal{L} = \text{SmoothL1Loss}(\hat S_y, \text{SG}(S_y))
\end{equation}
where $\text{SG}(\cdot)$ denotes the stop-gradient operation that prevents gradient flow through the teacher.

After updating the student and the predictor using the gradient of the loss, we update the teacher using an exponential moving average of the student's parameters.

\subsubsection{Training Efficiency}

Training on variable resolutions offers two significant advantages. First, it improves convergence speed and helps models generalize to resolutions unseen during training \cite{dehghani2023patchnpacknavit, bolya2025perceptionencoderbestvisual}. Second, it substantially increases sample throughput.

Our parallel data loading pipeline, inspired by Patch N' Pack, allows workers to resize, pack, and patch images concurrently. With a nominal batch size of $256$, our approach effectively processes approximately $1,200$ images per batch and $2,000$ images per second, increasing image throughput by a factor of $4\times$\footnote{We compare image throughput by training an identical model using the original IJEPA code. We reduce the batch size to 224 to avoid OOM.  The original IJEPA achieves (224 images per batch / 0.45 seconds per batch) = 500 images per second. }. Unlike the original IJEPA, where the sequence length of the batch supplied to the student and teacher varies between steps, our packed batch has a static sequence length. Our method can make the most of $\texttt{torch.compile}$ because we do not need to compile the computational graph with dynamic sizes. Our method allows a machine equipped with a single RTX3090 GPU to train a 236m parameter ViT-small and 52m parameter predictor at a training throughput of 2,000 images per second. Compare this to the throughput of the original IJEPA on the same machine and model, which can train at only 500 images per second.

This efficiency gain makes state-of-the-art self-supervised visual representation learning accessible for researchers with limited computational resources. Additionally, sequence packing allows us to have mask rates that differ across different samples in a mini-batch, whereas in the original IJEPA the mask rate is fixed for all samples in a minibatch. 

\section{Experiments}

We train two small IJEPA models, one using layer normalization at the output of the encoder, and one with DynTanh at the output of the encoder, denoted as IJEPA-LN, and IJEPA-Tanh respectively.
IJEPA-Tanh simply replaces the LN in Equation \ref{eq:LN-postproc} with a DynTanh nonlinear activation. We use DynTanh without affine parameters, squashing features into the range [-1, 1] \cite{zhu2025transformersnormalization}.

\begin{equation}\label{eq:Tanh-postproc}
S_x = \text{tanh}(a_{student} \tilde S_x), S_y = \text{tanh}(a_{teacher}\tilde S_y)
\end{equation}

Where $a_{student}$ is a learnable vector, and $a_{teacher}$ is a slow moving EMA copy of the student's vector.

We train both models for 600 epochs on Imagenet1K. We compute the self supervised losses over 8,000 validation images and collect and analyze the distribution of all unreduced loss values.

\section{Results}

\begin{figure}
    \centering
    \includegraphics[width=0.4\linewidth]{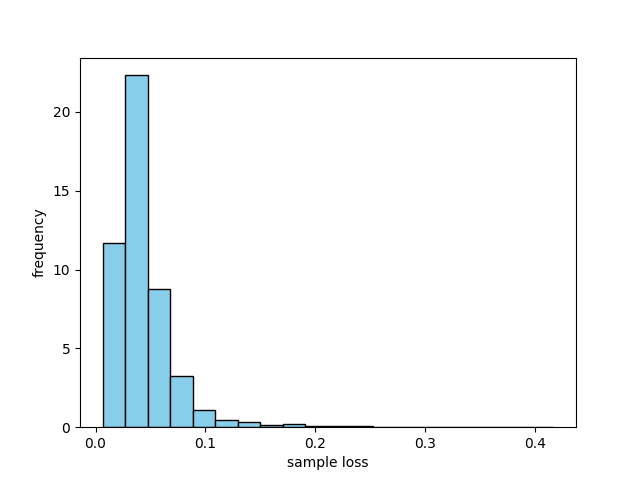}
    \includegraphics[width=0.4\linewidth]{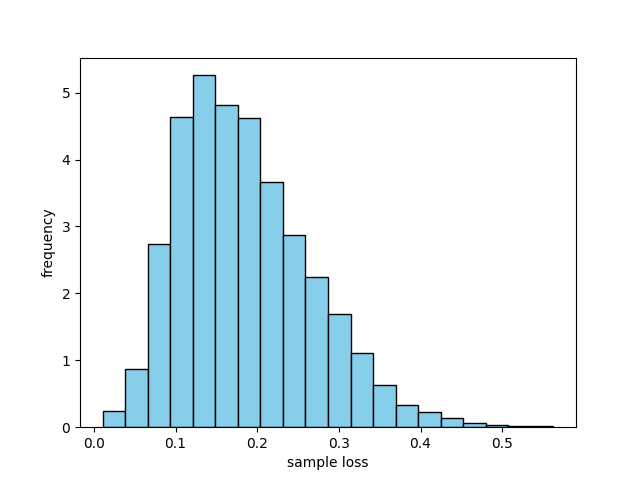}
    \caption{Left and right plots show the loss distributions of IJEPA-Tanh and IJEPA-LN respectively. Losses are measured over 8,000 validation images. IJEPA-Tanh exhibits a longer tail distribution, whereas IJEPA-LN more resembles a shorter tailed distribution. }
    \label{fig:ijepa-val-loss-distribution}
\end{figure}

We observe in Figure \ref{fig:ijepa-val-loss-distribution} that IJEPA-Tanh exhibits a longer tailed loss distribution than IJEPA-LN. We believe that the regularization introduced by DynTanh introduces a synergy where high energy tokens are more surprising and thus exhibit higher loss scores. IJEPA-LN on the other hand, cannot greatly accentuate differences in semantic importance, because the activations of every token are normalized to the standard normal distribution. 

\begin{figure}
    \centering
    \includegraphics[width=0.85\linewidth]{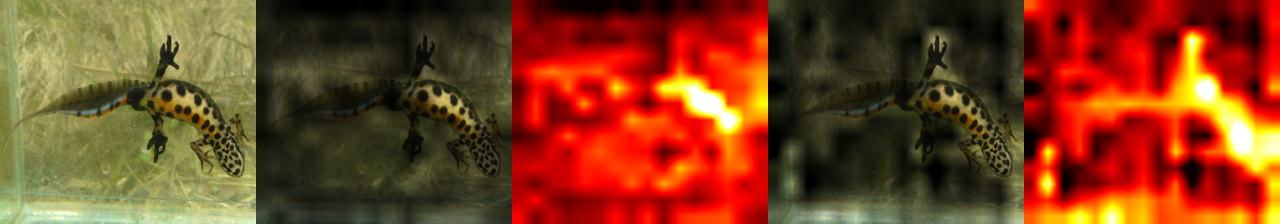}
    \includegraphics[width=0.85\linewidth]{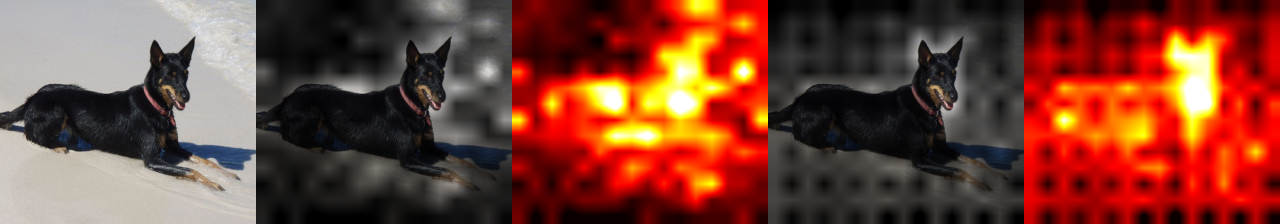}
    \includegraphics[width=0.85\linewidth]{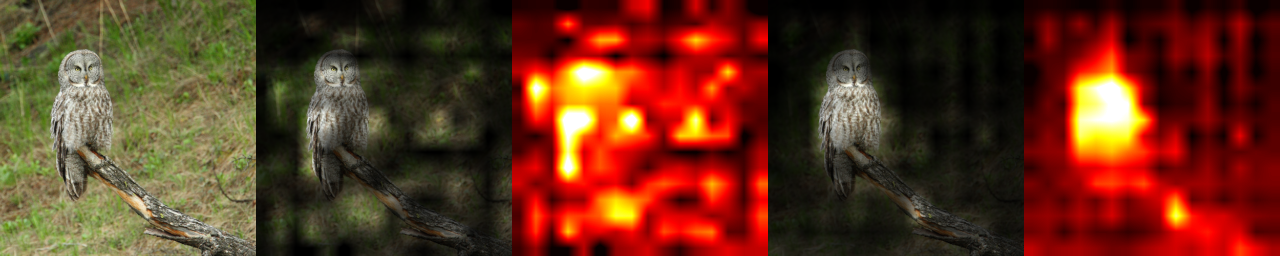}
    \includegraphics[width=0.85\linewidth]{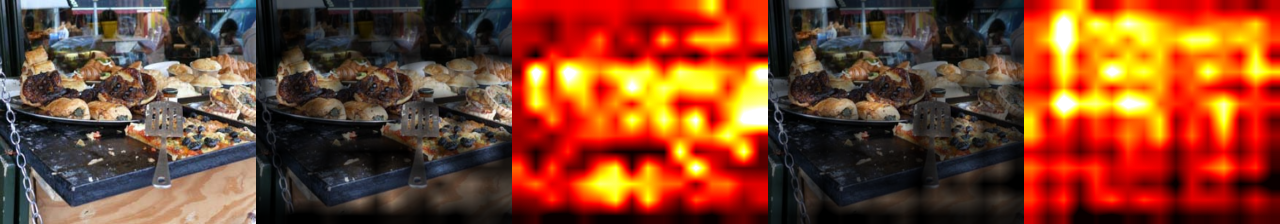}
    \includegraphics[width=0.85\linewidth]{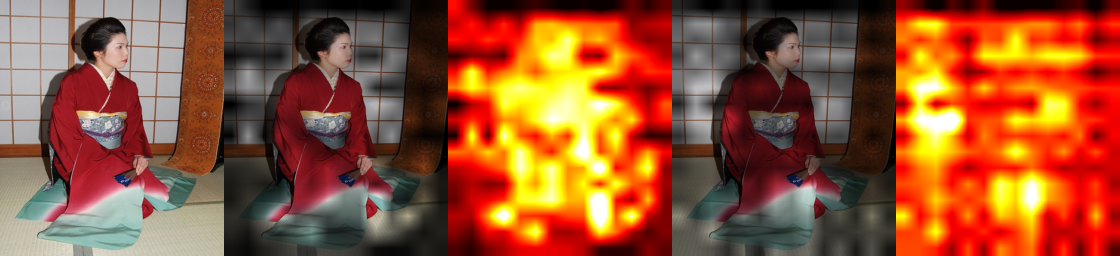}
    \caption{IJEPA losses computed from 8,000 random masks, each row showing a different image. For each row from left to right: Original image, multiplied loss map of IJEPA-Tanh, loss heatmap of IJEPA-Tanh, multiplied loss map of IJEPA-LN, loss heatmap of IJEPA-LN.}
    \label{fig:loss-images}
\end{figure}

To better compare IJEPA-Tanh and IJEPA-LN we visualize the spatial distribution of the self supervised losses. Figure \ref{fig:loss-images} visualizes the self supervised losses for several images, averaged over 8,000 different random context-target masks. This reveals the regions in the image that have higher losses. As expected, for both IJEPA-Tanh and IJEPA-LN, the foreground has the highest losses. 

\begin{figure}
    \centering
    \includegraphics[width=0.3\linewidth]{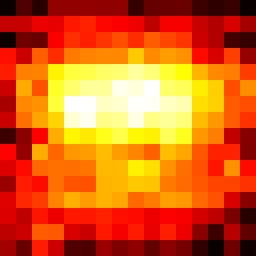}
    \includegraphics[width=0.3\linewidth]{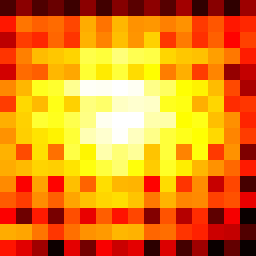}
    \caption{Loss heatmaps averaged over 6,500 samples. Left, IJEPA-Tanh, right: IJEPA-LN. The checkerboard artifact can be seen in IJEPA-LN's loss heatmap. }
    \label{fig:loss-heatmaps-mean}
\end{figure}

We observe an interesting pattern - the loss maps of IJEPA-LN exhibit a checkerboard of high losses. This presents as a grid of high losses persisting without regard to the semantic content in the image. Perhaps even more intriguing is that this checkerboard pattern is better seen when looking at losses averaged across many samples. We compute IJEPA losses across 6,500 samples, computing a mean loss image for both IJEPA-Tanh and IJEPA-LN. We average the losses to produce a smoothed loss map. As seen in Figure \ref{fig:loss-heatmaps-mean}, the checkerboard artifact in IJEPA-LN contrasts starkly with the smooth distribution of losses in IJEPA-Tanh.

We measure the validation performance of the two models on classification and monocular depth estimation. We observe that IJEPA-Tanh obtains consistent gains in validation performance. We encode features from the training set of Imagenet1k, and encode them using the teacher encoder. We mean pool the features across the sequence dimension to obtain a single vector embedding per image. We train a linear probe for 50 epochs, then evaluating the performance on the held-out validation set. We find that IJEPA-Tanh boosts the Imagenet1K classification accuracy of a linear probe from 38.021\% to 42.71\%. 

We measure the quality of both model's features for the dense prediction task of monocular depth estimation. We train a DPT-head\cite{ranftl2021visiontransformersdenseprediction} on the test set of NYU-depth-V2 for 10 epochs, then testing the performance on the validation set. We find that IJEPA-Tanh decreases the root squared mean error on NYU-Depth-v2 from 0.6273 to 0.6163 compared to IJEPA-LN.

\section{Limitations}

We find that simply replacing the encoder's penultimate LN with DynTanh results in smoother loss maps and better downstream performance. However, IJEPA-Tanh still uses LN in the intermediate transformer layers. While we would have liked to test further modifications, our current test setup requires training the model fully for several hundred epochs to only then reveal the differences in capabilities. We believe that future work remains to improve the intermediate insights into the features that the model is learning. Compared to MAE, which directly predicts pixel values, IJEPA models learn features that are difficult to comprehend. We hope for future IJEPA architectures that are natively trained to produce human-interpretable visualizations that encompass the thoughts and reflections that the model is making. 

\section{Conclusion}

We demonstrate the importance of preserving the relative energies of features learned by IJEPA models. Our key finding is that the standard LN applied to encoder features disrupts the natural energy hierarchy of visual tokens, where high-energy tokens encode semantically important image regions. By replacing layer normalization with DynTanh, we preserve these energy relationships while maintaining training stability.

Our empirical results show that this simple modification yields consistent improvements across diverse tasks: ImageNet linear probe accuracy increases from 38\% to 42.7\% for ViT-Small, and monocular depth estimation RMSE decreases by 0.08 on NYU Depth V2. Notably, IJEPA-Tanh exhibits a longer-tailed loss distribution and eliminates the checkerboard artifacts present in loss maps of IJEPA-LN models, suggesting that the model learns to focus on semantically meaningful regions rather than being constrained by artificially imposed uniformity.

These findings suggest that feature normalization strategies in self-supervised learning deserve more careful consideration. While normalization techniques like LN provide training stability, they may inadvertently constrain the model's ability to learn hierarchical representations. Our work opens avenues for future research into energy preserving architectures that preserve semantic structure while maintaining optimization stability. The consistent improvements across classification and dense prediction indicate that preserving natural token energies is a fundamental principle for effective self-supervised visual representation learning.

\bibliographystyle{plain}
\bibliography{sample}

\newpage
\section{Supplementary Material}

\subsection{Comparison to Previous Works}

BYOL \cite{grill2020bootstraplatentnewapproach} ablates the usage of L2 normalization at the output of the projector. They show that removing L2 normalization decreases Imagenet linear probe accuracy from 72.5\% to 67.4\%. Replacing L2 normalization with batch normalization decreases the accuracy further to 65.3\%. (The relevant experiment can be found in Section F.6, Table 20). Similarly, iBOT's distribution centering is critical for learning high quality representations, without which the linear probe accuracy drops from 74.2\% to 63.5\%. One notable characteristic is apparent across different works; all successful postprocessing methods use token-wise operations to project each token to a regularized space.

\subsection{Training and Architecture}

We use a modified ViT architecture for the encoder and the predictor. Instead of normalizing the input RGB values along the channel dimension, we simply scale them to be in the range [-1,1]. We use a patch size of 16 and use RoPe2D for position information. We use DiffMOE as a drop in replacement for the MLP in the transformer block \cite{shi2025diffmoedynamictokenselection}. Both reported models (IJEPA-Tanh and IJEPA-LN) use 16 experts. We use QK normalization in all attention layers. The architecture and training hyperparameters of IJEPA-Tanh and IJEPA-LN are identical, except that we replace the LN at the end of the encoder with a DynTanH layer. 

We train both ViT-small models on a single machine at a batch size of 256. We start with a learning rate of 1e-4, increasing it linearly for 10,000 warmup steps until it reaches 5e-4, then keeping it constant for the remainder of training. We use the AdamW optimizer with betas of (0.9,0.95) and weight decay of 0.05. We warmup the EMA $\beta$ value for the teacher. We start it at 0.95, linearly increasing for 1,000 steps until it reaches 0.999, then linearly increasing it for 300,000 steps until it reaches 0.9995, where it remains constant for the remainder of training. 

We train on a preprocessed training set of Imagenet1k \footnote{\href{https://huggingface.co/datasets/adams-story/imagenet1k-256-wds}{https://huggingface.co/datasets/adams-story/imagenet1k-256-wds}}. We resize images that have a side length larger than 256 so that their new maximum side length is 256. Images with a maximum side length smaller than 256 are not resized. This preserves the aspect ratio of the original image while significantly increasing the dataloading speed.

\begin{figure}
    \centering
    \includegraphics[width=0.5\linewidth]{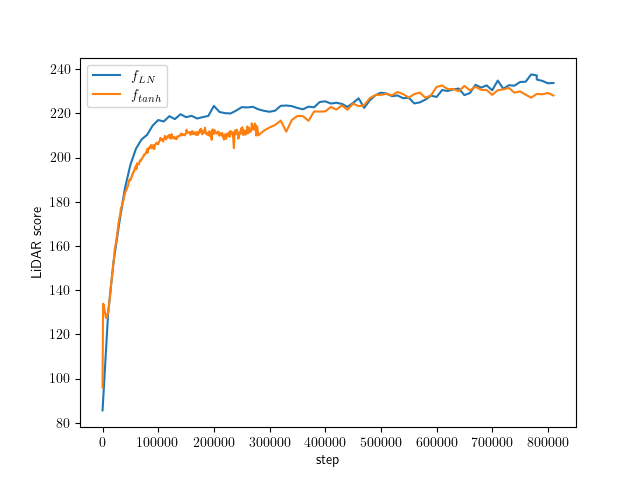}
    \caption{LiDAR score \cite{thilak2023lidarsensinglinearprobing} measured during pretraining.}
    \centering
    \includegraphics[width=0.5\linewidth]{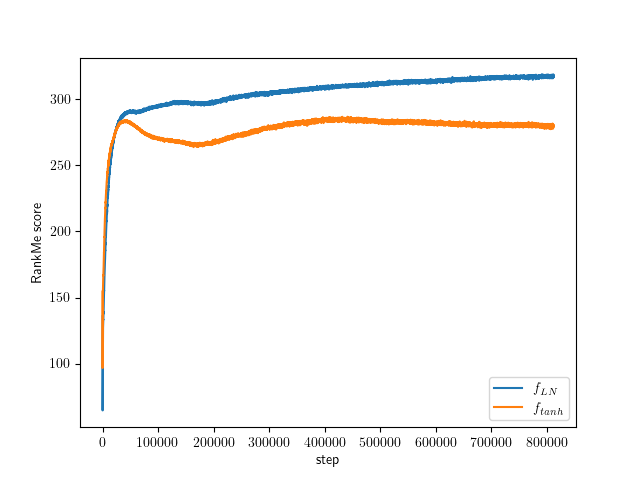}
    \caption{RankMe score \cite{thilak2023lidarsensinglinearprobing} measured during pretraining.}
    \centering
    \includegraphics[width=0.5\linewidth]{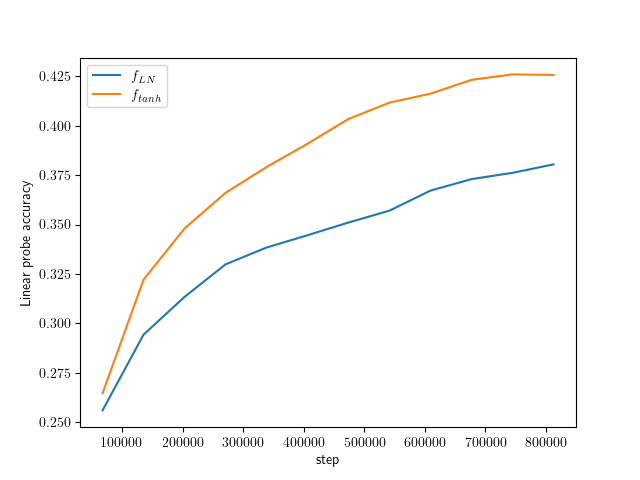}
    \caption{Linear probe accuracy measured during pretraining.}
    \label{fig:step-vs-oracle-scores}
\end{figure}

We record the RankMe score \cite{garrido2023rankmeassessingdownstreamperformance} and LiDAR score \cite{thilak2023lidarsensinglinearprobing} during pretraining. The different curves can be seen in Figure \ref{fig:step-vs-oracle-scores}. Both RankMe and LiDAR proved to be poor oracles, with IJEPA-Tanh exhibiting smaller RankMe and LiDAR scores than IJEPA-LN. LiDAR and RankMe scores plateaued after 150,000 training steps despite both models showing continued improvements in linear probe accuracy past this point.  

\begin{figure}
    \centering
    \includegraphics[width=0.5\linewidth]{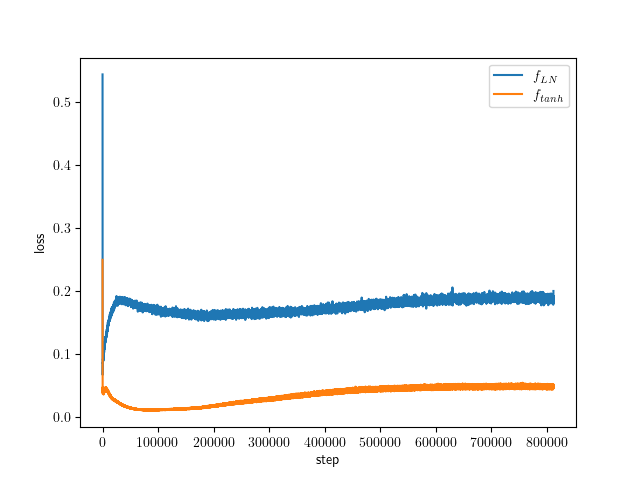}
    \caption{Loss during pretraining.}
    \label{fig:step-vs-loss}
\end{figure}

We observed that the model's training dynamics are sensitive to the shard shuffling strategy. Our first implementation shuffled shards without resampling. In this setup, each image is seen exactly once at a random point in a training epoch. When we resumed training from a checkpoint saved in the middle of an epoch we observed a loss spike and a drop in the RankMe score. This is probably due to the model overfitting to the order of images observed during an epoch. We employed a simple fix: simply shuffling shards with replacement fixed the loss spike when resuming training. All reported models were trained with shuffling with replacement.

\subsection{Linear Classification}

\begin{figure}
    \centering
    \includegraphics[width=0.5\linewidth]{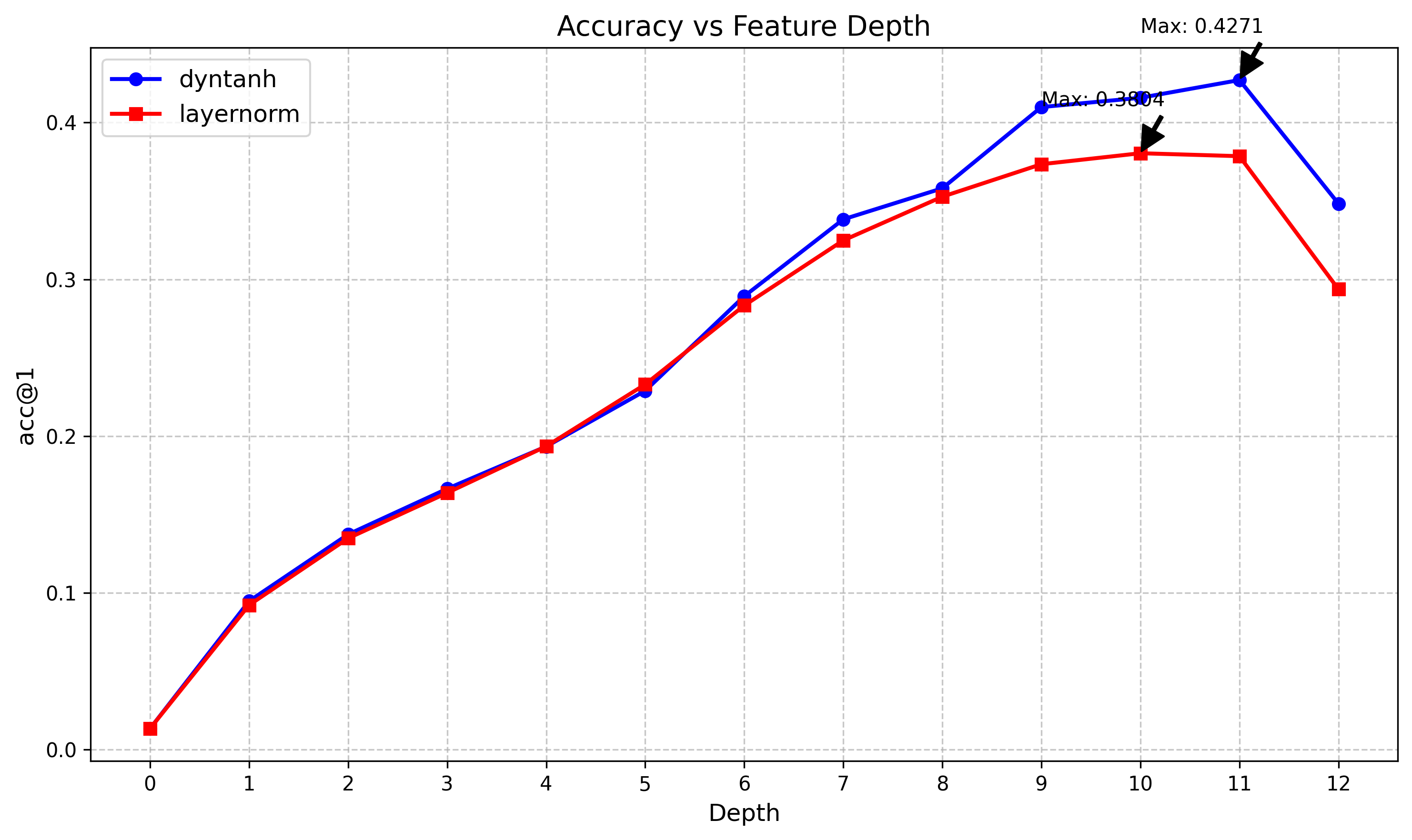}
    \caption{Linear probe accuracy on Imagenet1k across mean-pooled features extracted from different layers of the teacher encoder. }
    \label{fig:depth-vs-acc}
\end{figure}

We evaluate the discriminative performance of the encoder by training a linear probe to predict the labels of Imagenet. We do not use any data augmentation, simply square cropping all images to the minimum of their original (H,W), and then resizing to the a constant resolution of (256,256). We extract the hidden states from each layer of the encoder.  Notably, we do not apply the final LN or DynTanh to the features from the last layer, we found that using the encoder's final LN or DynTanh did not yeild better linear classification performance. We mean pool the hidden states along the sequence dimension. We train the probe on the imagenet training split with a constant learning rate of 1e-3 and a batch size of 2048 for 50 epochs, using the Adam optimizer. We then evaluate the accuracy on the validation set and report the maximum accuracy across all feature depths. Results can be seen in Figure \ref{fig:depth-vs-acc}. We observed that the performance degraded significantly at the last layer of the encoder, peaking somewhere between the fourth to last and second to last layer.

\subsection{Monocular Depth Estimation}

While our models are good at learning highly discriminative features, we also evaluate them on the more granular task of depth prediction. We train a DPT-head \cite{ranftl2021visiontransformersdenseprediction} over frozen encoder features from the third to last layer. The DPT-head produces a scale and shift invariant depth map, which is compared to a ground truth depth map that has been scaled and shifted. We train the DPT-head for 10 epochs on the training split of NYU Depth V2 \cite{Silberman:ECCV12:nyudepthv2} using the losses defined in \cite{ranftl2020robustmonoculardepthestimation}. For validation, we scale and and shift the model's predicted depth map using the scale and shift computed from the ground truth depth map. We report the RMSE between the model's scaled and shifted depth map and the raw ground truth depth map averaged over all samples in the validation split of NYU Depth V2. 

\begin{figure}
    \centering
    \includegraphics[width=0.6\linewidth]{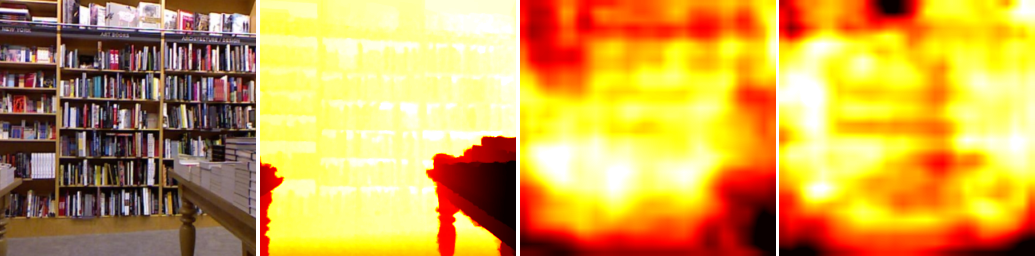}
    \includegraphics[width=0.6\linewidth]{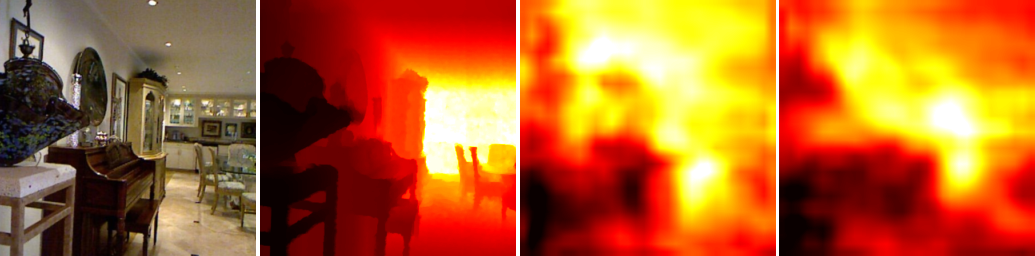}
    \includegraphics[width=0.6\linewidth]{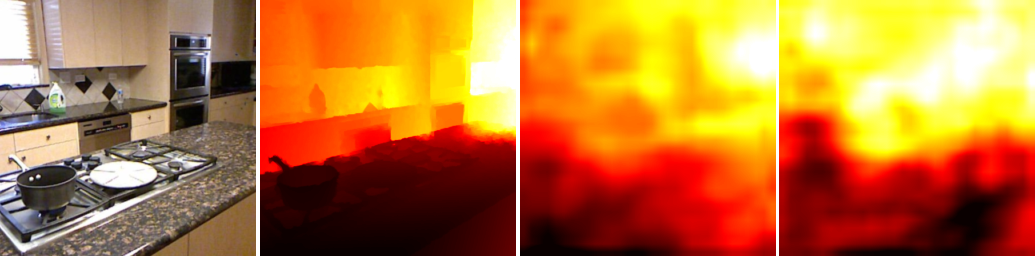}  
    \includegraphics[width=0.6\linewidth]{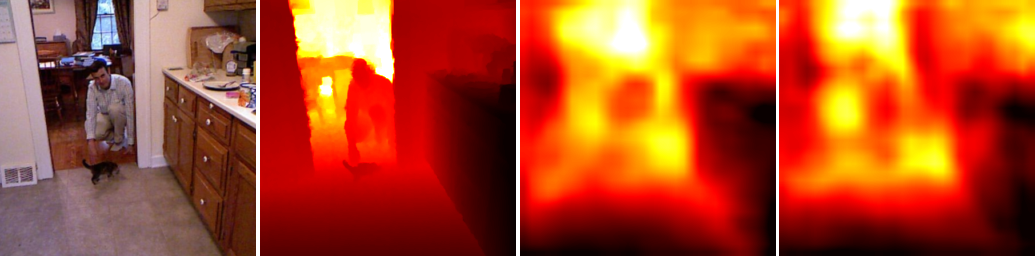}
    \includegraphics[width=0.6\linewidth]{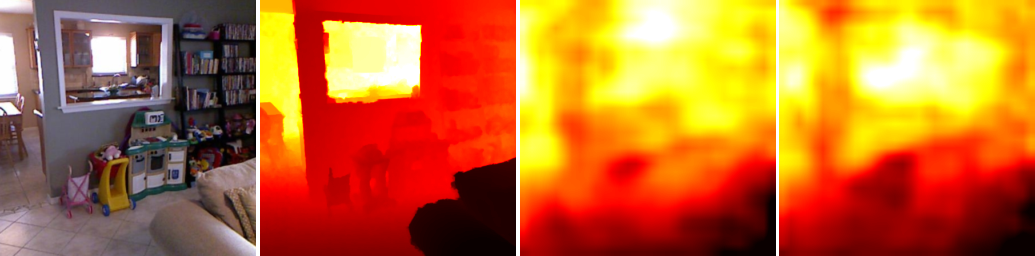} 
    \caption{Depth estimation outputs for several validation samples from NYU-Depth-V2. From the left to right: original image, ground truth depth map, IJEPA-Tanh depth map, IJEPA-LN depth map}
    \label{fig:depth-estimation-outputs}
\end{figure}

We visualize our model's predicted depth map for samples from the validation split of NYU Depth V2. We scale depth maps to fit in the range [0,1], and use the 'hot' colormap. Results can be seen in Figure \ref{fig:depth-estimation-outputs}.  

\subsection{Additional Results}

In addition to the two main models reported in the main body of this paper, we report several other training runs with different configurations in Table \ref{tab:additional-results}. Notably, across all of these configurations, the models trained with DynTanh at the end of the encoder perform better than models trained with LN at the end of the encoder. Some models we train we evaluate without the use of dynamic experts, where the capacity predictor \cite{shi2025diffmoedynamictokenselection} is discarded.  

We show the strength of our pretraining regimen by comparing a 100 epoch training run of our DynTanh model, to a 100 epoch training run using the code from IJEPA. We train identical models apart from replacing the final LN with a DynTanh. Our method drastically increases linear probe accuracy from 12.578\% to 32.216\%.

\rowcolors{2}{gray!10}{white}

\begin{table}[htbp]
\centering
\scriptsize
\setlength{\tabcolsep}{3pt}
\begin{tabular}{|p{0.8cm}|p{1.2cm}|p{1cm}|p{1.6cm}|p{0.5cm}|p{0.8cm}|p{0.7cm}|p{0.8cm}|p{0.75cm}|p{0.9cm}|p{0.6cm}|p{0.7cm}|p{1.2cm}|p{1.0cm}|}
\hline
\textbf{Method} & \textbf{Feature Norm} & \textbf{Samples per Batch} & \textbf{EMA Schedule} & \textbf{LR} & \textbf{LR Cool-down} & \textbf{Grad Scaler} & \textbf{Dynamic Experts} & \textbf{Experts} & \textbf{Mask Window (px)} & \textbf{Epoch} & \textbf{Wall Time (h)} & \textbf{ImageNet Acc@1} & \textbf{Depth RMSE}\cite{Silberman:ECCV12:nyudepthv2} \\
\hline
Ours & DynTanh & 1200 & Constant 0.996 & 5e-4 & no & no & no & 8 & 32 & 280 & 40 & \textbf{39.956\%} & 0.61886 \\
Ours & LayerNorm & 1200 & Constant 0.996 & 5e-4 & no & no & no & 8 & 32 & 280 & 40 & 34.61\% & \textbf{0.60804} \\
\hline
Ours & DynTanh & 1200 & Increasing 0.996$\to$0.9999 & 1e-3 & 40k & yes & no & 8 & 64 & 350 & 50 & \textbf{39.49\%} & 0.63489 \\
Ours & LayerNorm & 1200 & Increasing 0.996$\to$0.9999 & 1e-3 & 40k & yes & no & 8 & 64 & 350 & 50 & 37.588\% & \textbf{0.61457} \\
\hline
Ours & DynTanh & 950 & Increasing 0.999$\to$0.9995 & 5e-4 & no & yes & yes & 16 & 32 & 600 & 116 & \textbf{42.71\%} & 0.6163 \\
Ours & LayerNorm & 950 & Increasing 0.999$\to$0.9995 & 5e-4 & no & yes & yes & 16 & 32 & 600 & 116 & 38.021\% & \textbf{0.6273} \\
\hline
Ours & DynTanh & 470 & Increasing 0.999$\to$0.9995 & 5e-4 & no & yes & yes & 32 & 32 & 600 & 172 & \textbf{45.08\%} & - \\
\hline
Ours & DynTanh & 950 & Increasing 0.999$\to$0.9995 & 5e-4 & no & yes & yes & 16 & 32 & 100 & 20 & \textbf{32.216\%} & - \\
IJEPA\cite{assran2023selfsupervisedlearningimagesjointembedding} & LayerNorm & 224 & Increasing 0.995$\to$0.9999 & 5e-4 & no & yes & yes & 16 & Na & 100 & 73 & 12.578\% & - \\
\hline
\end{tabular}
\caption{Additional experimental results}
\label{tab:additional-results}
\end{table}

\subsection{Feature Visualization}

We use a similar technique as \cite{bolya2025perceptionencoderbestvisual} to visualize the encoder's features. We resize images to a random side length between 256 and 512 before encoding them. We smooth the features using a gaussian blur kernel. Then we take the top three PCA components. We then upscale feature patches using bilinear interpolation to increase the spatial size of the features to the input resolution. We treat these top three PCA components as hue, saturation, and value. We scale these values to fit in the range [0,1] before converting them to RGB pixels. 

\begin{figure}
    \centering
    \includegraphics[width=0.6\linewidth]{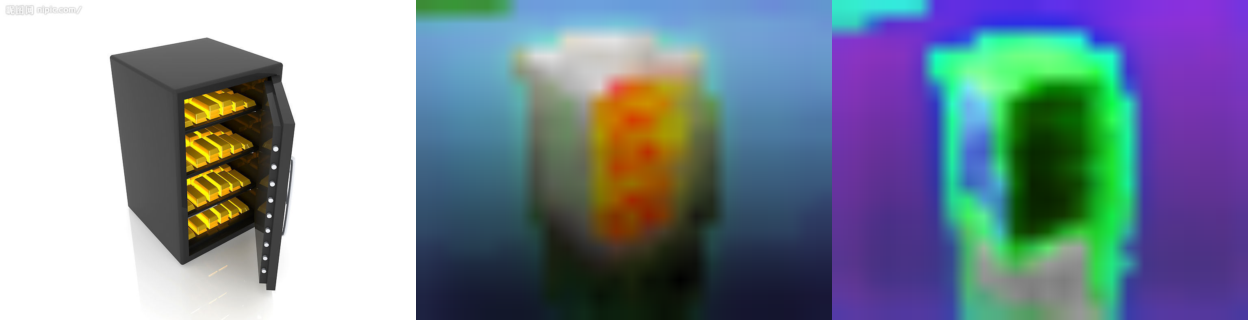}
    \includegraphics[width=0.6\linewidth]{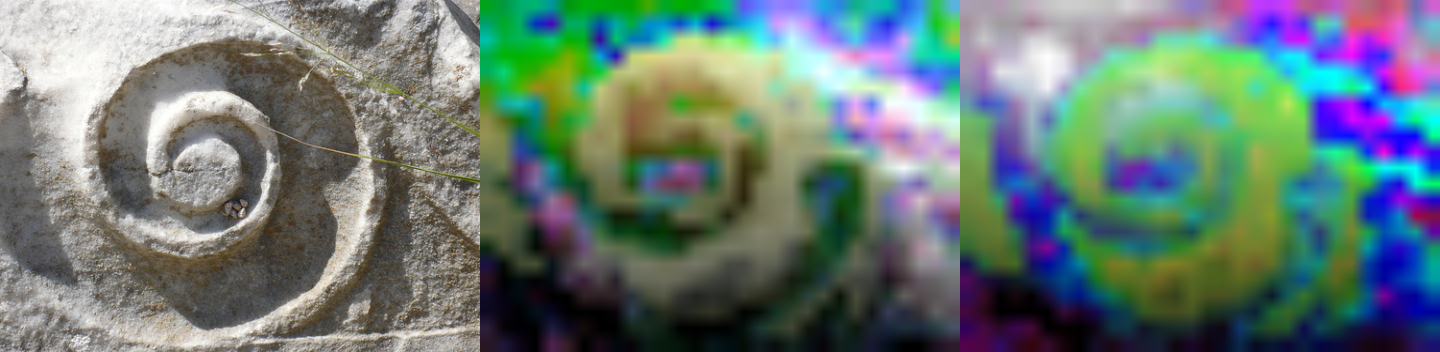}
    \includegraphics[width=0.6\linewidth]{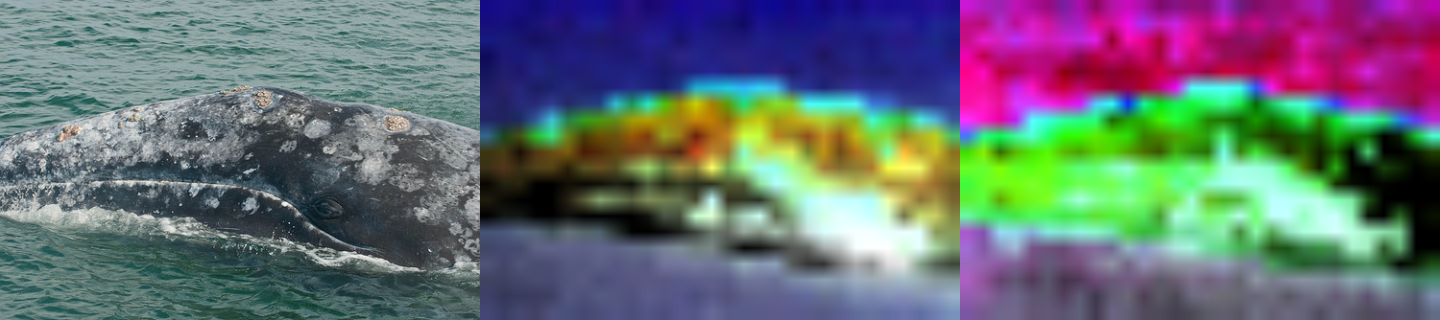}
    \includegraphics[width=0.6\linewidth]{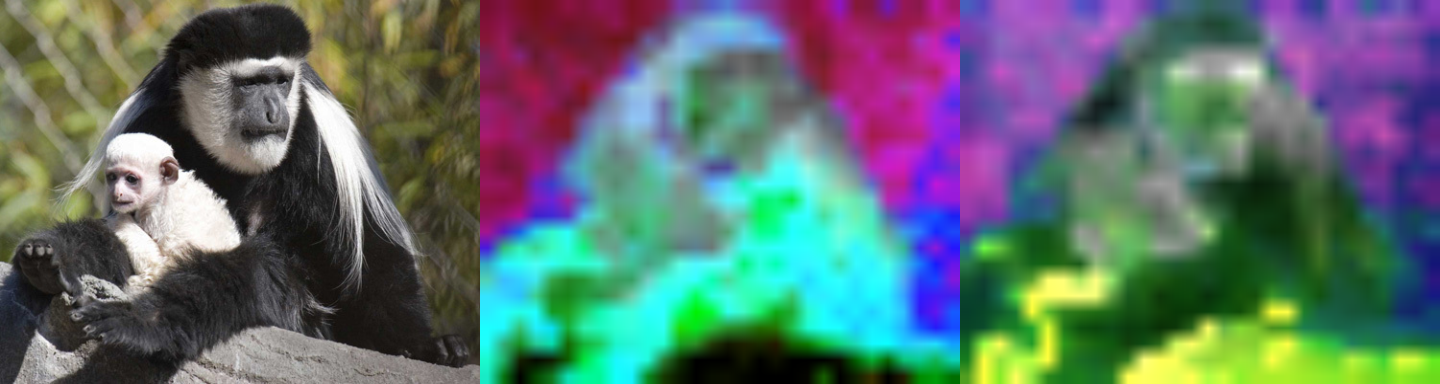}
    \includegraphics[width=0.6\linewidth]{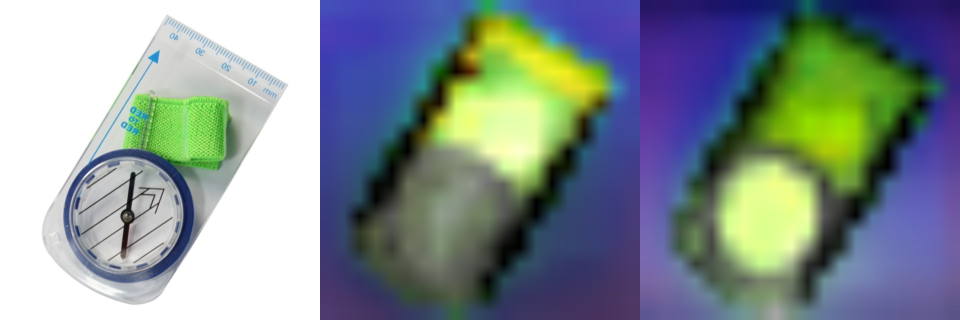}
    \includegraphics[width=0.6\linewidth]{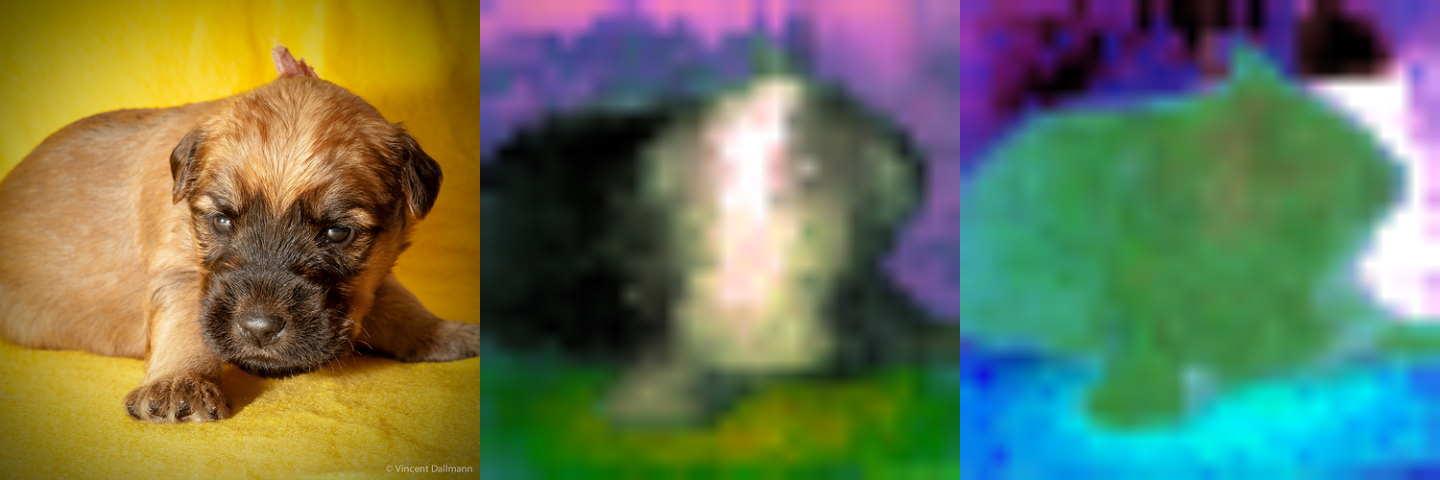}
    \includegraphics[width=0.6\linewidth]{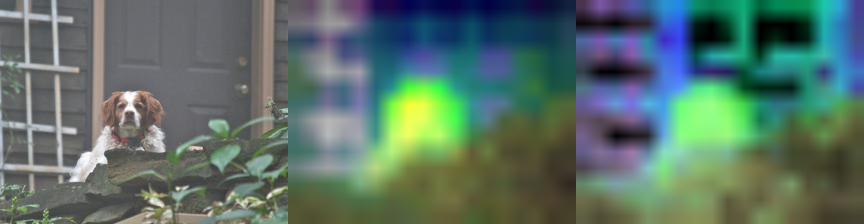}
    \includegraphics[width=0.6\linewidth]{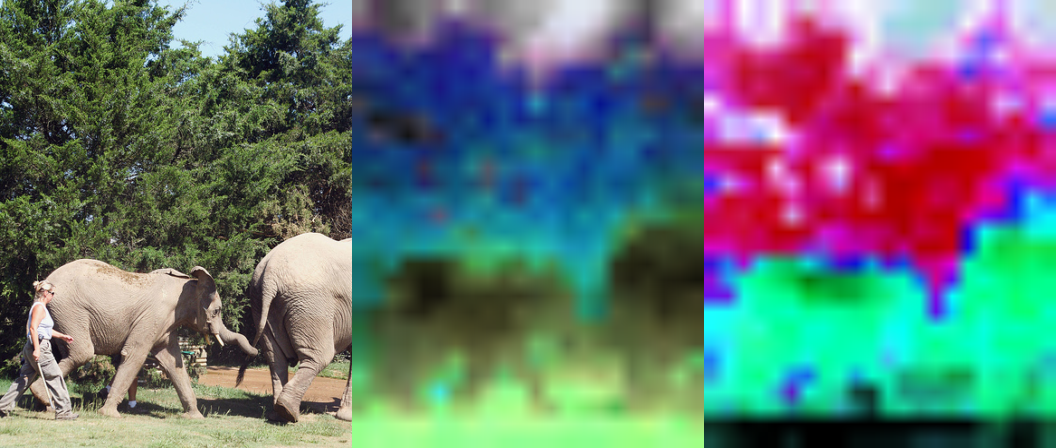}
    \caption{From left to right, original image, IJEPA-Tanh features, IJEPA-LN features}
    \label{fig:embeddings-visuals}
\end{figure}

\end{document}